\documentclass[final,3p]{elsarticle}

\makeatletter
\def\ps@pprintTitle{%
  \let\@oddhead\@empty
  \let\@evenhead\@empty
  \def\@oddfoot{Published in Neurocomputing (https://doi.org/10.1016/j.neucom.2012.07.034)\hfil}
  \let\@evenfoot\@oddfoot
}
\makeatother
    
\usepackage{graphicx}

\usepackage{amssymb}

\usepackage[caption=false,font=footnotesize]{subfig}
\usepackage[ruled,noline]{algorithm2e}

\biboptions{sort&compress}

\newcommand{\bbbr}{\mathbb{R}}

\journal{Neurocomputing}

\begin{document}

\begin{frontmatter}



\title{Reliable Probabilistic Classification with Neural Networks}


\author{Harris Papadopoulos}
\ead{h.papadopoulos@frederick.ac.cy}

\address{Frederick Research Center, 7-9 Filokyprou St.,\\Palouriotisa, Nicosia 1036, Cyprus}
\address{Computer Science and Engineering Department, Frederick University,\\7 Y. Frederickou St., Palouriotisa, Nicosia 1036, Cyprus}

\begin{abstract}
Venn Prediction (VP) is a new machine learning framework for producing 
well-calibrated probabilistic predictions. In particular it provides well-calibrated 
lower and upper bounds for the conditional probability of an example belonging to 
each possible class of the problem at hand. This paper proposes five 
VP methods based on Neural Networks (NNs), which is one of the most 
widely used machine learning techniques. The proposed methods are 
evaluated experimentally on four benchmark datasets and the obtained 
results demonstrate the empirical well-calibratedness of their outputs and 
their superiority over the outputs of the traditional NN classifier.
\end{abstract}


\end{frontmatter}


\section{Introduction}\label{sec:intro}

Machine learning techniques are becoming increasingly popular for solving all 
kinds of problems that cannot be solved with conventional tools. They have been 
applied to a great variety of problems and fields with very good results. 
However, most machine learning techniques do not provide any indication 
about the uncertainty of each of their predictions, which would have been 
very beneficial for most applications and especially for risk sensitive settings 
such as medical diagnosis \cite{holst:lack}. An indication of the likelihood of each 
prediction being correct notifies the user of a system about how much he can rely
on each prediction and enables him to take more informed decisions.

A solution to this problem was given by a recently developed machine learning 
theory called \emph{Conformal Prediction} (CP)~\cite{vovk:alrw}. CP can be used for extending traditional 
machine learning algorithms and developing methods (called Conformal Predictors) 
whose predictions are guaranteed to satisfy a given level of confidence without 
assuming anything more than that the data are independently and identically distributed (i.i.d.). 
More specifically, CPs produce as their predictions a set containing all the 
possible classifications needed to satisfy the required confidence level. To date many
different CPs have been developed, see e.g. 
\cite{lambrou:gacp,nouret:tcm-rr,papa:icpnn,papa:nnetricp,papa:icm-rr,papa:jairnnr,proedrou:tcm-pr,saunders:twcc,bhatt:rfcp}, 
and have been applied 
successfully to a variety of important problems such as the early detection of ovarian cancer~\cite{gam:preprot}, 
the classification of leukaemia subtypes~\cite{belloti:qualified},
the recognition of hypoxia electroencephalograms (EEGs)~\citep{zhang:hypoxia},
the recognition of gestures~\cite{ida:gesture},
the prediction of plant promoters~\cite{gam:plantprom},
the diagnosis of acute abdominal pain~\cite{papa:eisaap},
the assessment of the risk of complications following a coronary drug eluting stent procedure~\cite{vineeth:cardio},
the assessment of stroke risk~\cite{lambrou:stroke} and the
estimation of effort for software projects~\cite{papa:soft}.
The CP framework has also been extended to additional problem settings such as 
active learning~\cite{ho:active} and change detection in data streams~\cite{ho:streamchange}.

This paper focuses on an extension of the original CP framework, called Venn 
Prediction (VP), which can be used for making \emph{multiprobability predictions}. 
In particular multiprobability predictions are a set of probability distributions
for the true classification of the new example. This set can be summarized by lower 
and upper bounds for the conditional probability of the new example belonging to 
each one of the possible classes\footnote{It should be noted that moving from multiprobability 
predictions to the corresponding lower and upper bounds entails some loss of information,
since the set of probability predictions for each class is replaced by their maximum 
and minimum values.}. The resulting bounds are guaranteed to contain 
well-calibrated probabilities (up to statistical fluctuations).
Again, like with CPs, the only assumption made for obtaining this guaranty is 
that the data are i.i.d.

The VP framework has until now been combined with the \emph{k}-nearest neighbours 
algorithm in \cite{vovk:alrw} and \cite{dash:vennit} and with Support Vector Machines 
in \cite{zhou:vennplatts}. A Venn Predictor based on Neural Networks (NNs) was first 
proposed in \cite{papa:nnvpaiai} for binary 
classification problems. This work is extended here by developing five Venn 
Predictors based on NNs for multiclass problems, for which NNs have more than 
one output neurons. The choice of NNs as basis for the proposed methods was made 
for two main reasons: (a) their popularity among machine learning techniques for 
almost any type of application, 
see e.g. \cite{anasta:annprognosis,mantza:aiaiosteop,pattichis:annmis,iliadis:water,harala:broadcasting,iliadis:woodinfsci},
and (b) that they can produce probabilistic outputs which can be compared with 
those produced by the Venn Predictors. The experiments performed examine 
on one hand the empirical well-calibratedness of the probability bounds produced by the 
proposed methods and on the other hand compare them with the probabilistic outputs
of the traditional NN classifier. 

The rest of this paper starts with an overview of the Venn Prediction 
framework in the next section, while in Section~\ref{sec:NNVP} it 
details the proposed Neural Network Venn Prediction methods.
Section~\ref{sec:Res} presents the experiments performed on four 
benchmark datasets and reports the obtained results. Finally, Section~\ref{sec:Conc} 
gives the conclusions and future directions of this work.

\section{The Venn Prediction Framework}\label{sec:Venn}

This section gives a brief description of the Venn prediction framework; for
more details the interested reader is referred to \cite{vovk:alrw}. We are 
given a training set $\{(x_1, y_1), \dots, (x_l, y_l)\}$ of examples\footnote{The ``training set'' 
is in fact a multiset, as it can contain some examples more than once.}, where 
each $x_i \in \bbbr^d$ is the vector of attributes for example $i$ and 
$y_i \in \{Y_1, \dots, Y_c\}$ is the classification label of that example. We are 
also given a new unclassified example $x_{l+1}$ and our task is to predict 
the probability of this new example belonging to each class 
$Y_j \in \{Y_1, \dots, Y_c\}$ based only on the assumption that 
all $(x_i, y_i), i= 1, 2, \dots$ are generated independently by the same 
probability distribution (i.i.d.).

The main idea behind Venn prediction is to divide all examples into a 
number of categories based on their similarity and calculate the 
probability of $x_{l+1}$ belonging 
to each class $Y_j \in \{Y_1, \dots, Y_c\}$ as the frequency of $Y_j$
in the category that contains it. However, as we don't know the true 
class of $x_{l+1}$, we assign each one of the possible classification labels 
to it in turn and for each assigned classification label $Y_k$ we calculate a 
probability distribution for the true class of $x_{l+1}$ based on
the examples
\begin{equation}
\label{eq:extset}
  \{(x_1, y_1), \dots, (x_l, y_l), (x_{l+1}, Y_k)\}.
\end{equation}

To divide each set (\ref{eq:extset}) into categories we use what we call 
a \emph{Venn taxonomy}. A Venn taxonomy defines a number of categories and a 
rule based on which each example is assigned to one of these categories; 
formally a category is a multiset of examples. In effect similar examples 
should be assigned to the same category so that the resulting probability 
distribution for each assumed classification label $Y_k$ will
depend on the examples that are most similar to $(x_{l+1}, Y_k)$.

Typically each taxonomy is based on a traditional
machine learning algorithm, called the \emph{underlying algorithm} of the 
Venn predictor. The output of this algorithm for each attribute vector 
$x_i, i = 1,\dots, l+1$ after being trained either on the whole set (\ref{eq:extset}), 
or on the set resulting after removing the pair $(x_i, y_i)$ from (\ref{eq:extset}),
is used to assign $(x_i, y_i)$ to one of a predefined set of categories.
For example, a Venn taxonomy that can be used with every traditional algorithm 
puts in the same category all examples that are assigned the same 
classification label by the underlying algorithm. At this point it is important to 
emphasize the difference between the classes of the problem and the categories of a 
Venn taxonomy. Even though this example taxonomy consists of a category corresponding 
to each classification label, these categories are assigned examples based 
on the output classification label of the underlying algorithm and not on the 
true class to which each example belongs. Therefore the category corresponding to 
a given classification label $Y_k$ will contain the examples that the underlying 
algorithm ``believes'' to belong to class $Y_k$, which are not necessarily the same 
as the examples that actually do belong to that class since the underlying algorithm 
might be wrong in some cases. Of course other Venn taxonomies can be defined that 
depend on more information obtained from the underlying algorithm rather than just 
the output classification label. Four new Venn taxonomies for multiclass Neural Networks 
are defined in the next section.

After partitioning (\ref{eq:extset}) into categories using a Venn taxonomy,
the category $T_{new}$ containing the new example $(x_{l+1}, Y_k)$ will be 
nonempty as it will contain at least this one example. Then the empirical 
probability of each classification label $Y_j$ in this category will be
\begin{equation}
\label{eq:prob}
  p^{Y_k}(Y_j) = \frac{|\{(x^*, y^*) \in T_{new} : y^* = Y_j\}|}{|T_{new}|}.
\end{equation}
This is a probability distribution for the label of $x_{l+1}$. After 
assigning all possible classification labels to $x_{l+1}$ we get a set of probability
distributions that compose the multiprobability prediction of the 
Venn predictor $P_{l+1} = \{p^{Y_k} : Y_k \in \{Y_1, \dots, Y_c\}\}$.
As proved in \cite{vovk:alrw} the predictions produced by any Venn predictor are 
automatically well-calibrated multiprobability predictions. This is true regardless 
of the taxonomy of the Venn predictor. Of course the taxonomy used is 
still very important as it determines how efficient, or informative, the 
resulting predictions are. We want the diameter of multiprobability 
predictions and therefore their uncertainty to be small, since saying that 
the probability of a given classification label for an example is 
between $0.8$ and $0.9$ is much more informative than saying that it 
is between $0$ and $0.9$. We also want the predictions to be as close 
as possible to zero or one, indicating that a classification label is 
highly unlikely or highly likely respectively.

The maximum and minimum probabilities obtained for each classification label $Y_j$ 
define the interval for the probability of the new example belonging 
to $Y_j$:
\begin{equation}
\label{eq:interval}
\bigg[\min_{k = 1,\dots,c} p^{Y_k}(Y_j), \max_{k = 1,\dots,c} p^{Y_k}(Y_j)\bigg].
\end{equation}
To simplify notation the lower bound of this interval for a given class $Y_j$ will 
be denoted as $L(Y_j)$ and the upper bound will be denoted as $U(Y_j)$.
The Venn predictor outputs the class $\hat y = Y_{j_{best}}$ as its prediction where
\begin{equation}
\label{eq:pred}
j_{best} = \arg\max_{j = 1,\dots, c} \overline{p(Y_j)},
\end{equation}
and $\overline{p(Y_j)}$ is the mean of the probabilities obtained for $Y_j$:
\begin{equation}
\label{eq:mean}
\overline{p(Y_j)} = \frac{1}{c} \sum_{k = 1}^{c} p^{Y_k}(Y_j).
\end{equation}
This prediction is accompanied by the interval 
\begin{equation}
\label{eq:PredInterval}
[L(\hat y), U(\hat y)]
\end{equation}
as the probability interval of it being correct. The complementary 
interval 
\begin{equation}
\label{eq:ErrInterval}
[1 - U(\hat y), 1 - L(\hat y)]
\end{equation}
gives the probability that $\hat y$ is not the true classification label of 
the new example and it is called the \emph{error probability interval}.

\section{Venn Prediction with Neural Networks}\label{sec:NNVP}

This section describes the proposed Neural Network based Venn Prediction 
methods. The NNs used were fully connected feed-forward networks with a single 
hidden layer consisting of units with a hyperbolic tangent activation function.
Their output layer consisted of units with a softmax activation function~\cite{bridle:softmax},
which made all their outputs lie between zero and one and their sum equal to one.
They were trained with the scaled conjugate gradient algorithm~\cite{moller:scg} 
minimizing cross-entropy error (log loss):
\begin{equation}\label{eq:ce}
CE = -\sum^{N}_{i=1} \sum^{c}_{j=1} t^j_i\log(o^j_i),
\end{equation}
where $N$ is the number of examples, $c$ is the number of possible classes, 
$o^1_i, \dots, o^c_i$ are the outputs of the network for example $i$ and 
$t^1_i, \dots, t^c_i$ is the binary form of the true classification 
label $y_i$ of example $i$, that is
\begin{equation}\label{eq:binrep}
t^j_i = \left\{
\begin{array}{ll}
1 , & \textrm{if $y_i = Y_j$,} \\
0 , & \textrm{otherwise.}
\end{array}\right.
\end{equation}
As a result the outputs of these NNs can be interpreted as probabilities for 
each class and can be compared with those produced by the proposed methods.

As explained in Section~\ref{sec:Venn} the difference between alternative 
Venn Prediction methods is the taxonomy they use to divide examples
into categories. Here five different Venn taxonomies are defined which 
allocate examples into categories based on the outputs $o^1_i, \dots, o^c_i$ 
of the NN for each example $i$ after being trained on the 
extended set (\ref{eq:extset}). 

\begin{algorithm}
\KwIn{training set $\{(x_1, y_1), \dots, (x_l, y_l)\}$, new example $x_{l+1}$, possible classes $\{Y_1, \dots, Y_c\}$.}
\For{$k = 0$ \KwTo $c$}
{
   Train the NN on the extended set $\{(x_1, y_1), \dots, (x_l, y_l), (x_{l+1}, Y_k)\}$\;
   Supply the input patterns $x_1, \dots, x_{l+1}$ to the trained NN to obtain the outputs $o_1, \dots, o_{l+1}$\;
   \For{$i = 0$ \KwTo $l+1$}
   {
      Assign $(x_i, y_i)$ to the corresponding category $T_m$ of the Venn taxonomy $V_1, V_2, V_3, V_4$, or $V_5$ 
      according to the NN outputs $o^1_i, \dots, o^c_i$\;
   }
   Find the category $T_{new}$ that contains $(x_{l+1}, Y_k)$\;
   \For{$j = 0$ \KwTo $c$}
   {
      $p^{Y_k}(Y_j) := \frac{|\{(x^*, y^*) \in T_{new} : y^* = Y_j\}|}{|T_{new}|}$\;
   }
}
\For{$j = 0$ \KwTo $c$}
{
   $\overline{p(Y_j)} := \frac{1}{c} \sum_{k = 1}^{c} p^{Y_k}(Y_j)$\;
}
\KwOut\\
\Indp
\Indp
Prediction $\hat y = \arg\max_{j = 1, \dots, c} \overline{p(Y_j)}$\;
The probability interval for $\hat y$: $[\min_{k = 1, \dots, c} p^{k}(\hat y), \max_{k = 1, \dots, c} p^{k}(\hat y)]$.
\caption{Neural Networks Venn Predictor\label{al:nnvp}}
\end{algorithm}

The first Venn taxonomy, which will be denoted as $V_1$, is the simple 
taxonomy that assigns two examples to the same category if their maximum 
outputs correspond to the same class. This produces $c$ categories, one 
for each possible class of the problem. This is a taxonomy that can be used 
with any traditional classifier as underlying algorithm of the Venn Predictor.
The remaining four taxonomies defined below were developed in this work especially 
for being used with Neural Networks as the underlying algorithm of the 
Venn Predictor. These taxonomies take into account more information 
about the actual values of the NN outputs, rather than just which one is the 
maximum output, and divide examples more effectively into more than $c$ categories. 
Consequently the 
taxonomy categories are smaller than those of $V_1$ and consist of examples 
that are more similar to each other, resulting in more accurate probabilistic 
outputs.

The second taxonomy, which will be denoted as $V_2$, further divides 
the examples in each category of taxonomy $V_1$ into two smaller categories 
based on the value of their maximum output.
It is expected that the higher the value of the maximum output for an example, 
the higher the chance of the corresponding class being the correct one. 
Therefore the examples of each category of taxonomy $V_1$ are divided into 
those with maximum output above a high threshold $\theta$ and those 
with maximum output below $\theta$. This produces $2c$ categories. 
In principle $\theta$ can be set to any value between $\frac{1}{c}$ and $1$, 
which is the range of possible values for the maximum output of the NN, 
however values in the upper half of this range are most appropriate since 
typically the maximum outputs of NN are relatively high. Here $\theta$ is 
set to $0.75$ for all experiments with this taxonomy. 

The third taxonomy, which will be denoted as $V_3$, again further divides 
each category of taxonomy $V_1$ into two smaller categories, but this time 
the division depends on the value of the second highest 
output of the examples. It is expected that the higher the value of the second 
highest output of an example, which corresponds to the most likely classification
after the predicted one, the lower the chance of the class corresponding to 
its maximum output being the correct one. Therefore the examples of each 
category of taxonomy $V_1$ are divided into those with second highest 
output above a low threshold $\theta$ and those with second highest output 
below $\theta$. This again produces $2c$ categories. In this case $\theta$ can 
be set to any value between $0$ and $0.5$, as the second highest output cannot 
be higher than $0.5$. Here it is set to $0.25$ for all experiments with this 
taxonomy.

The fourth taxonomy, which will be denoted as $V_4$, again further divides 
each category of taxonomy $V_1$ in two, but this time the division depends 
on the difference between the highest and second highest outputs. In effect 
this difference takes into account both how high the maximum output of the 
example is and how low the second highest output is. The bigger this difference 
is for an example, the higher the chance of the class corresponding 
to its maximum output being the correct one. Therefore the examples of each 
category of taxonomy $V_1$ are divided into those with difference between 
its two highest outputs above a threshold $\theta$ and those with difference 
below $\theta$. Like $V_2$ and $V_3$ this taxonomy also consists of $2c$ 
categories. In this case $\theta$ can 
be set to any value between $0$ and $1$, however values in the middle of this 
range are most appropriate as very small or very big differences are very unusual. 
Here $\theta$ is set to $0.5$ for all experiments with this taxonomy.

The fifth and last taxonomy, which will be denoted as $V_5$, assigns two 
examples to the same category if their outputs that are above a given low 
threshold $\theta$ correspond to the same set of classes. In effect this 
taxonomy considers all outputs above $\theta$ as likely and puts the examples 
that have the same likely classifications into the same category. In principle 
this taxonomy consists of $2^c$ categories, but most of them are almost always 
empty as having more than $2$ outputs above $\theta$ is extremely unusual. 
In this case $\theta$ can be set to a value between $0$ and $0.5$, as a 
$\theta \geq 0.5$ would never have more than one outputs as likely. 
Here $\theta$ is set to $0.25$ for all experiments with this taxonomy. 

Using these taxonomies the examples are divided into categories for each assumed 
classification label $Y_k \in \{Y_1, \dots, Y_c\}$ of $x_{l+1}$ and the process 
described in Section~\ref{sec:Venn} is followed for calculating the outputs of the 
Neural Network Venn Predictor (NN-VP). Algorithm~\ref{al:nnvp} presents the complete 
NN-VP algorithm.

\section{Experiments and Results}\label{sec:Res}

Experiments were performed on four datasets from the 
UCI Machine Learning Repository \cite{data:uci}, which has been widely used as a source 
for benchmark datasets in testing machine learning algorithms:
\begin{itemize}
\item \textbf{Teaching Assistant Evaluation}, which is concerned with the evaluation 
of 151 Teaching Assistant (TA) assignments over three regular semesters and two summer 
semesters at the Statistics Department of the University of Wisconsin-Madison.
The 151 TA assignments are described by 5 attributes and are divided into 3 score 
classes of roughly equal size: low, medium, high.

\item \textbf{Glass Identification}, which is concerned with the identification of glass 
types based on their oxide content, motivated by criminological investigations. The dataset 
consists of 214 glasses of 6 different types (classes). The number of glasses of each type 
ranges from $76$ to $9$. Each glass is described by 9 attributes.

\item \textbf{Ecoli}, which is concerned with the prediction of the localization sites 
of proteins. It consists of 336 instances described by 7 attributes and divided 
into 8 classes. The number of examples in each class ranges from $143$ to just $2$ and 
$91\%$ of the examples belong to $4$ out of the $8$ classes.

\item \textbf{Vehicle Silhouettes}, which was gathered at the Turing Institute, Glasgow, Scotland. 
It is concerned with the classification of vehicle silhouettes into 4 types of vehicles. 
There are 846 instances, each described by 18 attributes extracted from the corresponding 
silhouette. 
\end{itemize}
In an effort to make the evaluation and comparison presented here as general as possible the 
four datasets used correspond to different problem domains. It is also worth to note that 
two of the datasets have a class imbalance problem, which results in the corresponding tasks 
being somewhat more difficult.

\begin{table}[b]
  \small
  \centering
  \begin{tabular}{@{\extracolsep{0.1cm}}rcccc} \hline\noalign{\smallskip}
                    & TA         & Glass          &       & Vehicle     \\
                    & Evaluation & Identification & Ecoli & Silhouettes \\ \noalign{\smallskip}\hline\noalign{\smallskip}
  Examples          & 151        & 214            & 336   & 846 \\
  Attributes        & 5          & 9              & 7     & 18  \\
  Classes           & 3          & 6              & 8     & 4   \\
  Hidden Neurons    & 5          & 5              & 10    & 11  \\\noalign{\smallskip}\hline
  \end{tabular}
  \caption{Main characteristics and number of hidden neurons used for each data set.}
  \label{tab:datasets}
\end{table}

The NNs used were fully connected feed-forward networks with a single hidden layer consisting of 
hyperbolic tangent units and an output layer consisting of softmax units. The number of hidden
units used for each dataset was experimentally selected by applying the traditional NN classifier 
on 10 random divisions of each dataset into training and test sets consisting of $90\%$ and $10\%$ 
of the examples respectively. It should be noted that these randomly generated sets are different 
from the ones used for evaluating the performance of the NN-VP in the batch setting described in 
Subsection~\ref{subsec:ResBatch}. Despite this difference there still remains an element of data 
snooping in the selection of the number of hidden units for each dataset, but this is in fact 
in favour of the traditional NN classifier, which is what the performance of NN-VP is compared 
against, since this was the classifier used for determining the number of hidden units. 
The selected number of hidden units for each dataset is 
reported in Table~\ref{tab:datasets} along with the dataset characteristics.

All NNs were trained with the scaled conjugate gradient algorithm~\cite{moller:scg} minimizing 
cross-entropy error~(\ref{eq:ce}) and early stopping based on a validation set consisting of $30\%$ of 
the corresponding training set. In an effort to avoid local minima each NN was trained $3$ times with different random
initial weight values and the one that performed best on the validation set was 
selected for being applied to the test examples. Before each training session all attributes 
were normalised setting their mean value to $0$ and their standard deviation to $1$. 

Two sets of experiments were performed: on-line experiments, presented in Subsection~\ref{subsec:ResOnline}, 
to demonstrate empirically that the probability bounds produced by NN-VP are well-calibrated and batch 
experiments, presented in Subsection~\ref{subsec:ResBatch}, to compare the performance of 
NN-VP with that of the traditional NN classifier and assess the performances of the five 
different Venn taxonomies.

\subsection{On-line Experiments}\label{subsec:ResOnline}

\begin{figure}
	\centering
		\subfloat[Teaching Assistant Evaluation]{\includegraphics[trim = 1mm 3mm 2mm 0mm, clip, width=6.8cm]{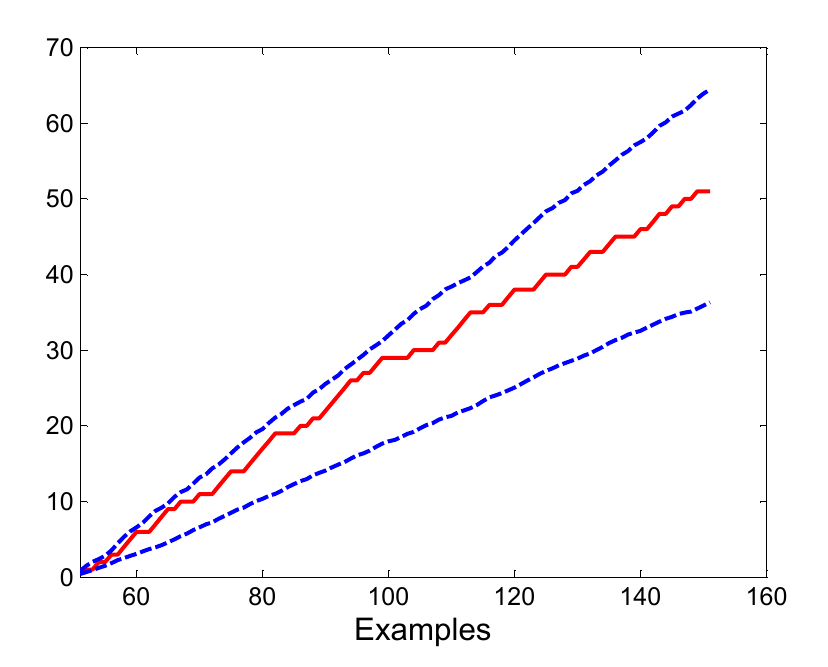}}
		\subfloat[Glass Identification]{\includegraphics[trim = 2mm 3mm 2mm 0mm, clip, width=6.8cm]{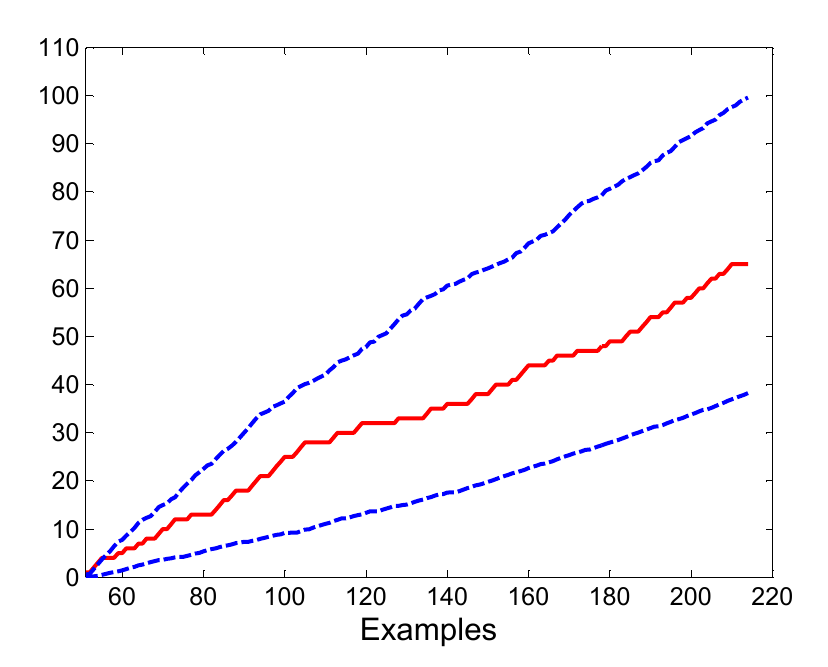}}\\
		\subfloat[Ecoli]{\includegraphics[trim = 2mm 3mm 2mm 0mm, clip, width=6.8cm]{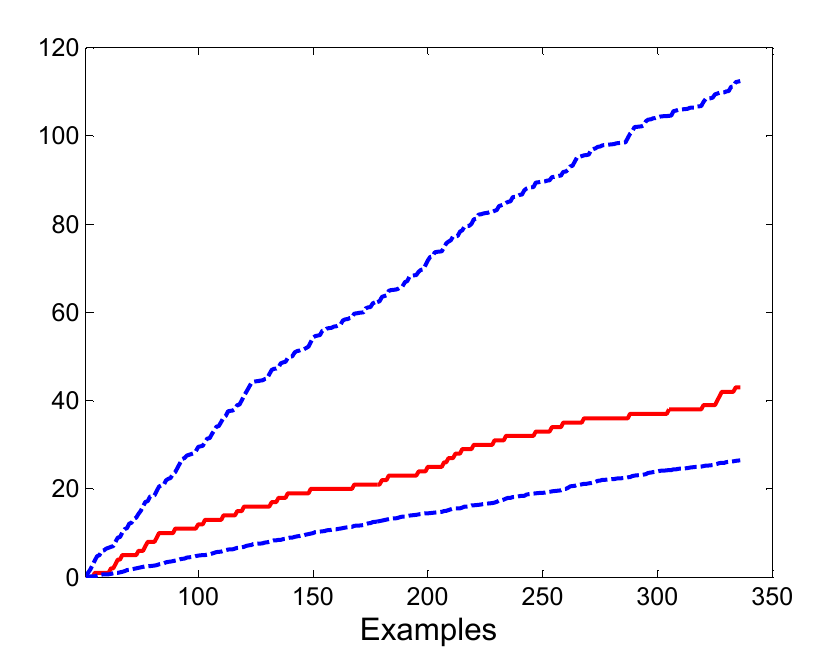}}
		\subfloat[Vehicle Silhouettes]{\includegraphics[trim = 2mm 3mm 2mm 0mm, clip, width=6.8cm]{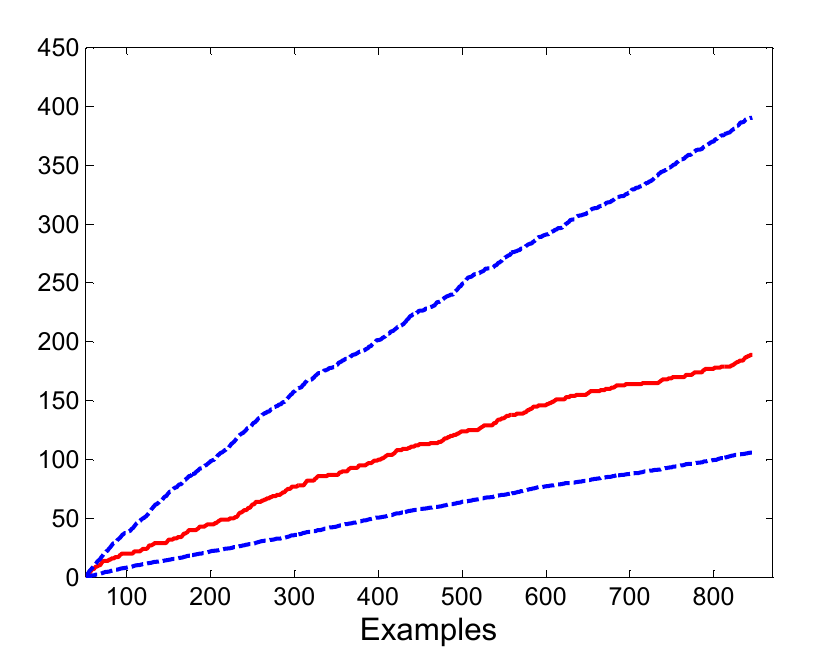}}
\caption{On-line performance of NN-VP with $V_1$ on the four datasets. Each plot shows the cumulative number of 
         errors $E_n$ with a solid line and the cumulative lower and upper error probability curves
         $LEP_n$ and $UEP_n$ with dashed lines.}
\label{fig:onlineVenn}
\end{figure}

\begin{figure}
	\centering
		\subfloat[Teaching Assistant Evaluation]{\includegraphics[trim = 1mm 3mm 2mm 0mm, clip, width=6.8cm]{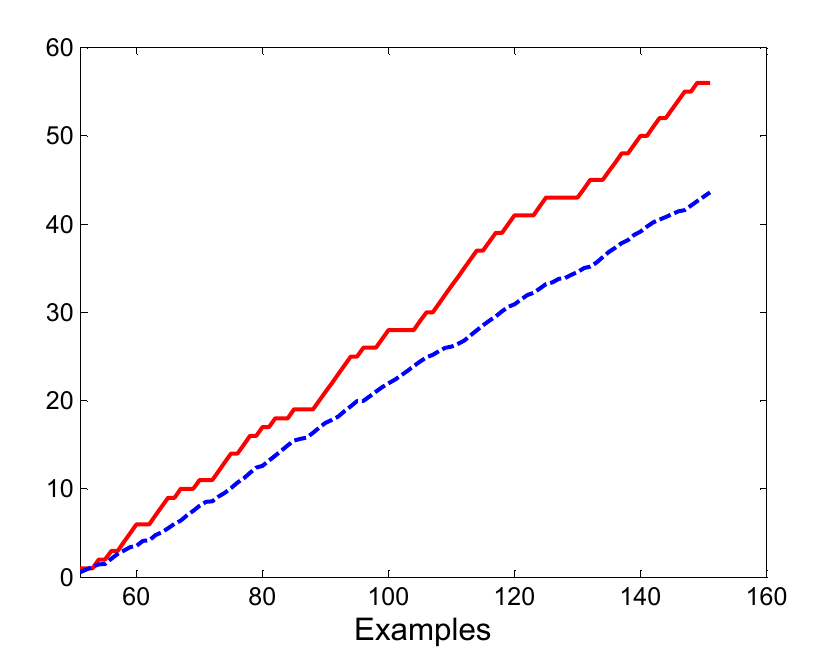}}
		\subfloat[Glass Identification]{\includegraphics[trim = 1mm 3mm 2mm 0mm, clip, width=6.8cm]{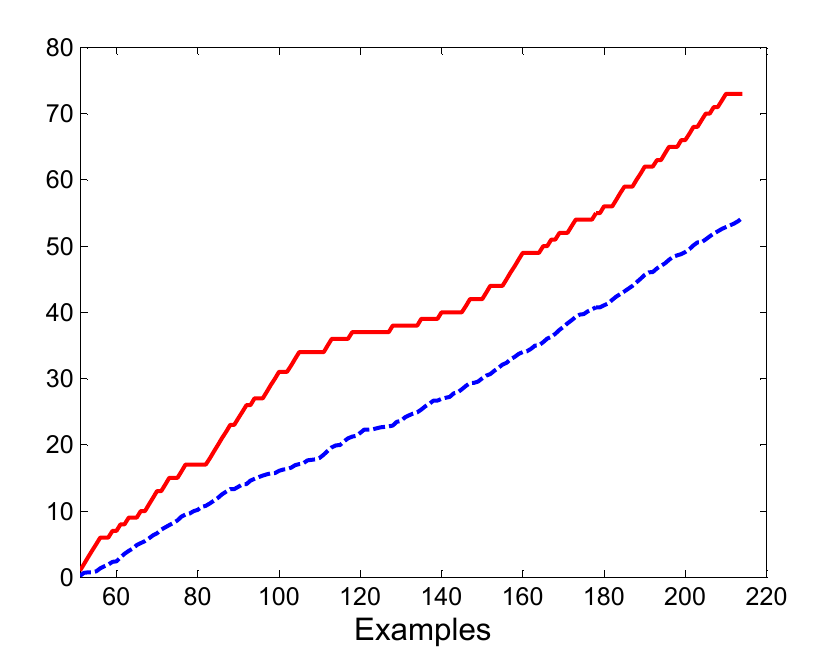}}\\
		\subfloat[Ecoli]{\includegraphics[trim = 1mm 3mm 2mm 0mm, clip, width=6.8cm]{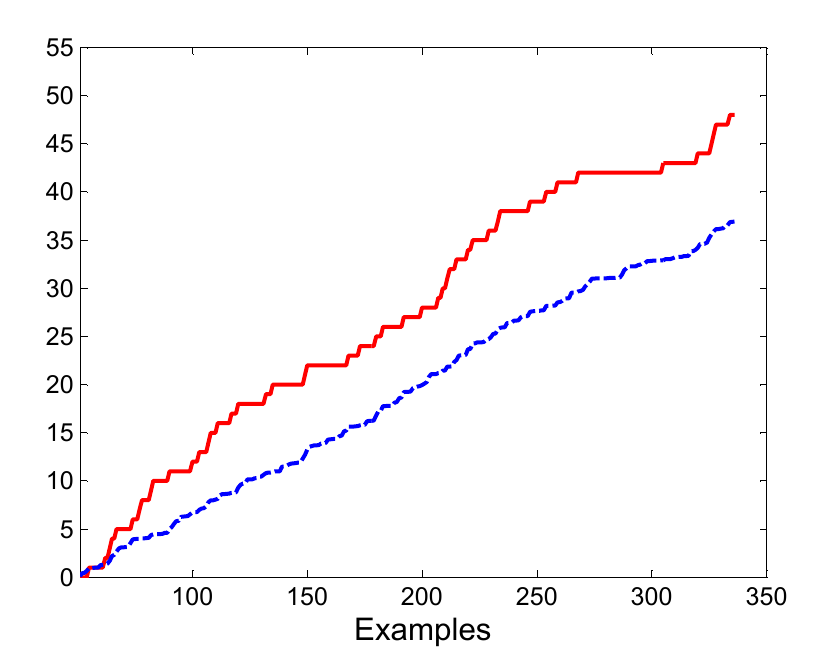}}
		\subfloat[Vehicle Silhouettes]{\includegraphics[trim = 1mm 3mm 2mm 0mm, clip, width=6.8cm]{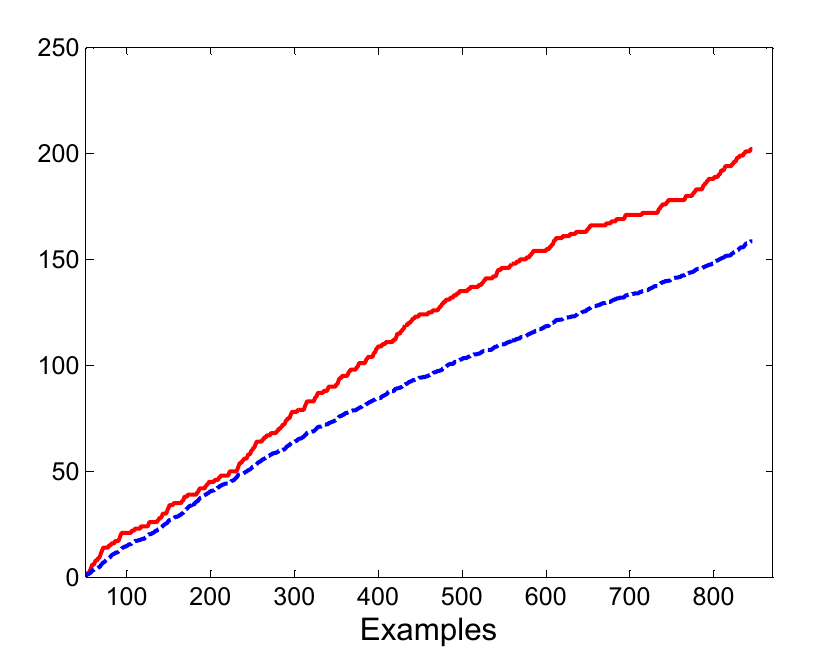}}
\caption{On-line performance of the traditional NN classifier on the four datasets. Each plot shows the 
         cumulative number of errors $E_n$ with a solid line and the cumulative error probability curve 
         $EP_n$ with a dashed line.}
\label{fig:onlineNN}
\end{figure}

This subsection demonstrates the empirical well-calibratedness of NN-VP 
by applying it to the four datasets 
in the on-line mode. More specifically, starting with an initial training 
set consisting of $50$ examples, each subsequent example is predicted in turn 
and then its true classification is revealed and it is added to the training set 
for predicting the next example. 

Figure \ref{fig:onlineVenn} shows the following 
three curves obtained by applying NN-VP with taxonomy $V_1$ on each dataset:
\begin{itemize}
\item the cumulative error curve
\begin{equation}
\label{eq:cumErr}
    E_n = \sum^n_{i=1} err_i,
\end{equation}
where $err_i = 1$ if the prediction $\hat y_i$ is wrong and $err_i = 0$ otherwise,
\item the cumulative lower error probability curve (see (\ref{eq:ErrInterval}))
\begin{equation}
\label{eq:cumLEP}
LEP_n = \sum^n_{i=1} (1 - U(\hat y_i))
\end{equation}
\item and the cumulative upper error probability curve
\begin{equation}
\label{eq:cumUEP}
UEP_n = \sum^n_{i=1} (1 - L(\hat y_i)).
\end{equation}
\end{itemize}
In effect the final cumulative errors are expected to lie between the final 
values of the cumulative upper and lower error probability curves up to statistical 
fluctuations. The four plots of Figure~\ref{fig:onlineVenn} confirm that 
the probability intervals produced by NN-VP are well-calibrated. 
In fact the cumulative errors are always included inside the cumulative 
upper and lower error probability curves produced by the NN-VP. The same is true 
for the plots obtained with the other four Venn taxonomies, which are very similar 
to the ones shown in this figure and are presented in~\ref{app1}. 

The analogous
plots generated by applying the traditional NN classifier to the four datasets are shown in 
Figure \ref{fig:onlineNN}. In this case the cumulative error curve (\ref{eq:cumErr}) 
for each NN is plotted together with the cumulative error probability curve
\begin{equation}
\label{eq:cumEP}
EP_n = \sum^n_{i=1} |1 - \hat p_i|,
\end{equation}
where $\hat p_i$ is the probability given by the NN for example $i$ belonging to 
its predicted classification $\hat y_i$:
\begin{equation}
\label{eq:phat}
\hat p_i = \max_{j = 1, \dots, c} o^j_i.
\end{equation}
In effect this curve is the sum of the probabilities of all classes except the predicted 
one for each example according to the NN. One would expect that this curve would be very near 
the cumulative error curve if the probabilities produced by the NN were well-calibrated. 
The plots of Figure \ref{fig:onlineNN} show that this is not the case. The NNs 
underestimate the true error probability in all cases since the cumulative error curve 
is much higher than the cumulative error probability curve. To check how misleading the 
probabilities produced by the NN are, the 2-sided p-value of obtaining a total number 
of errors $E_N$ with the observed deviation from the expected errors $EP_N$ given the 
probabilities produced by the NN was calculated for each dataset. 
The resulting p-values were $0.0087$ for the TA Evaluation dataset, $0.00091$ 
for the Glass Identification dataset, $0.026$ for the Ecoli dataset and $0.000013$ 
for the Vehicle Silhouettes dataset. The fact that three out of four of these p-values
are below the $0.01$ significance level demonstrates the need for probability intervals 
as opposed to single probability values as well as that the probabilities produced by 
NNs can be very misleading.

\subsection{Batch Experiments}\label{subsec:ResBatch}

This subsection examines the performance of NN-VP in the batch setting and 
compares its results with those of the direct predictions made by the traditional NN classifier.
For these experiments the four datasets were divided randomly into training 
and test sets consisting of $90\%$ and $10\%$ of their examples respectively. 
In order for the results not to depend on a particular division
into training and test sets, $10$ different random divisions were performed 
and all results reported here are over all $10$ test sets.

Since NNs produce a single probabilistic output $o^j$ for each possible classification $Y_j$, 
for NN-VP the values $\overline{p(Y_j)}$ corresponding to the estimate of NN-VP about 
the probability of each test example belonging to class $Y_j$ were used for this performance 
evaluation and comparison. Consequently in this Subsection $\hat o^j_i$ will be used to denote 
the probabilistic output of the method in question for classification $Y_j$ of example $i$, 
which in the case of the NN-VPs will correspond to $\overline{p_i(Y_j)}$.
For reporting these results four quality metrics are used. The first is the accuracy of 
each classifier, which does not take into account the probabilistic outputs produced, 
but it is the most popular metric for assessing the quality of classifiers. The second 
is cross-entropy error~(\ref{eq:ce}), which is in fact the error minimized by 
the training algorithm of the NNs on the training set. The third 
metric is the Brier score~\cite{brier:brscore}:
\begin{equation}
BS = \frac{1}{N} \sum^{N}_{i=1} \sum^{c}_{j=1} (\hat o^j_i - t^j_i)^2,
\end{equation}
where $t^1_i, \dots, t^c_i$ is the binary form of the true classification label $y_i$ (see~(\ref{eq:binrep})).
The cross-entropy error, or log-loss, and the Brier score are the 
most popular quality metrics for probability assessments.

The Brier score can be decomposed into three terms interpreted as 
the uncertainty, reliability and resolution of the probabilities, 
by dividing the range of probability values into a number of 
intervals $K$ and representing each interval $k = 1,\dots, K$ by a 
`typical' probability value $r_k$ \cite{murphy:brierpartition}.
The reliability term of this decomposition measures how close the output probabilities are 
to the true probabilities and therefore reflects how well-calibrated 
the output probabilities are. This is the most important component of 
interest in this work as it evaluates how much one can rely on the 
probabilistic outputs produced by each method. Hence this is the 
fourth metric used here. It is defined in \cite{murphy:brierpartition} as:
\begin{equation}
REL = \frac{1}{N} \sum^{K}_{k=1} n_k (r_k - \phi_k)^2,
\end{equation}
where $n_k$ is the number of $\hat o^j_i$, $i = 1, \dots, N$ and $j = 1, \dots, c$ 
in the interval $k$ and $\phi_k$ is the percentage of these outputs 
for which $t^j_i = 1$, i.e. the example belongs to the corresponding 
class of the output. Here the number of categories $K$ was set to $100$
for the first three datasets and to $200$ for the much larger Vehicle 
Silhouettes dataset.

\begin{table}[t]
  \centering
  \caption{Results of the traditional NN and the five NN-VPs on the TA Evaluation dataset}
  \label{tab:res1}
  \begin{tabular}{l @{\extracolsep{0.32cm}}r c c c c} \hline\noalign{\smallskip}
  Method          & & Accuracy & CE & BS & REL \\ 
  \noalign{\smallskip}\hline\noalign{\smallskip}
  \multicolumn{2}{l}{Traditional NN} & 45.33\% & 155.62 & 0.6323 & 0.0475 \\
  \noalign{\smallskip}\noalign{\smallskip}
                             & $V_1$ & 48.67\% & 156.27 & 0.6286 & 0.0283 \\
                             & $V_2$ & 47.33\% & 155.67 & 0.6249 & 0.0294 \\
  NN-VP                      & $V_3$ & 50.67\% & 152.98 & 0.6156 & 0.0335 \\
                             & $V_4$ & 48.67\% & 154.88 & 0.6230 & \textbf{0.0215} \\
                             & $V_5$ & \textbf{52.67\%} & \textbf{150.63} & \textbf{0.5997} & 0.0274 \\
  \noalign{\smallskip}\hline\noalign{\smallskip}
  \end{tabular}
\end{table}

\begin{table}[t]
  \centering
  \caption{Results of the traditional NN and the five NN-VPs on the Glass Identification dataset}
  \label{tab:res2}
  \begin{tabular}{l @{\extracolsep{0.32cm}}r c c c c} \hline\noalign{\smallskip}
  Method          & & Accuracy & CE & BS & REL \\ 
  \noalign{\smallskip}\hline\noalign{\smallskip}
  \multicolumn{2}{l}{Traditional NN} & 57.62\% & 236.37 & 0.5697 & 0.0172 \\
  \noalign{\smallskip}\noalign{\smallskip}
                             & $V_1$ & 58.57\% & 217.71 & 0.5354 & 0.0119 \\
                             & $V_2$ & 59.52\% & 216.00 & 0.5311 & 0.0124 \\
  NN-VP                      & $V_3$ & 60.00\% & 217.14 & 0.5306 & 0.0110 \\
                             & $V_4$ & \textbf{60.48\%} & \textbf{213.22} & \textbf{0.5245} & 0.0109 \\
                             & $V_5$ & 59.05\% & 216.35 & 0.5353 & \textbf{0.0089} \\
  \noalign{\smallskip}\hline\noalign{\smallskip}
  \end{tabular}
\end{table}

\begin{table}[t]
  \centering
  \caption{Results of the traditional NN and the five NN-VPs on the Ecoli dataset}
  \label{tab:res3}
  \begin{tabular}{l @{\extracolsep{0.32cm}}r c c c c} \hline\noalign{\smallskip}
  Method          & & Accuracy & CE & BS & REL \\ 
  \noalign{\smallskip}\hline\noalign{\smallskip}
  \multicolumn{2}{l}{Traditional NN} & 86.76\% & \textbf{151.30} & 0.2090 & 0.0061 \\
  \noalign{\smallskip}\noalign{\smallskip}
                             & $V_1$ & 88.53\% & 163.35 & 0.2023 & \textbf{0.0059} \\
                             & $V_2$ & 88.53\% & 164.36 & 0.2040 & 0.0060 \\
  NN-VP                      & $V_3$ & 89.12\% & 156.13 & \textbf{0.1948} & 0.0061 \\
                             & $V_4$ & 88.53\% & 161.52 & 0.2011 & 0.0061 \\
                             & $V_5$ & \textbf{89.41\%} & 156.61 & 0.1980 & \textbf{0.0059} \\
  \noalign{\smallskip}\hline\noalign{\smallskip}
  \end{tabular}
\end{table}

\begin{table}[t]
  \centering
  \caption{Results of the traditional NN and the five NN-VPs on the Vehicle Silhouettes dataset}
  \label{tab:res4}
  \begin{tabular}{l @{\extracolsep{0.32cm}}r c c c c} \hline\noalign{\smallskip}
  Method          & & Accuracy & CE & BS & REL \\ 
  \noalign{\smallskip}\hline\noalign{\smallskip}
  \multicolumn{2}{l}{Traditional NN} & 80.35\% & \textbf{356.56} & 0.2620 & 0.0139 \\
  \noalign{\smallskip}\noalign{\smallskip}
                             & $V_1$ & 80.82\% & 373.85 & 0.2567 & 0.0106 \\
                             & $V_2$ & 81.88\% & 371.81 & 0.2517 & \textbf{0.0086} \\
  NN-VP                      & $V_3$ & 81.65\% & 369.92 & 0.2516 & 0.0099 \\
                             & $V_4$ & \textbf{82.12\%} & 368.48 & \textbf{0.2494} & 0.0088 \\
                             & $V_5$ & 81.18\% & 380.11 & 0.2585 & 0.0106 \\
  \noalign{\smallskip}\hline\noalign{\smallskip}
  \end{tabular}
\end{table}

Tables \ref{tab:res1} to \ref{tab:res4} present the results of the traditional NN and the 
five NN-VPs on each dataset. The values in bold indicate the best performance for each 
metric. The values reported in these tables show that all five VPs perform better than the 
traditional NN classifier against all metrics except the cross-entropy error. However, even in the case 
of the cross-entropy error metric for two out of the four datasets the best values are obtained 
by one of the VPs. Overall, the differences between the traditional NN classifier and the five 
VPs in the values of the cross-entropy error and Brier score metrics are relatively small ranging 
from an increase of $8.6\%$ on the value of the traditional NN in the worst case to a decrease 
of $9.8\%$ in the best case. The differences in terms of accuracy are a bit more important 
ranging from an improvement of $0.6\%$ for the worst performing VP to an improvement of 
$16.2\%$ for the best performing VP. More significant differences are observed in the values 
of the reliability metric where for the TA Evaluation, Glass Identification and 
Vehicle Silhouettes datasets the improvement of VPs ranges from $23.7\%$ to $54.7\%$. 
This shows that even if we do not take into account the probabilistic intervals produced 
by the VPs and consider only the mean probabilities for each class, we can still achieve 
a considerable improvement over the reliability of the probabilities produced by the 
traditional NN classifier. In the case of the Ecoli dataset the reliability values of 
all methods are more or less the same probably because in this case the 
traditional NN classifier gives more reliable probabilities. Of course one cannot know 
if the same will be true for the particular problem and data he is working on. Moreover, 
even in such cases the probabilistic intervals of VPs are much more reliable since they 
are guaranteed to contain well-calibrated probabilities. 

When comparing the performance of the five Venn taxonomies between each other, 
one can see that in most cases the new taxonomies defined here, $V_2, V_3, V_4$ and $V_5$, 
perform better than the simple taxonomy $V_1$. In terms of accuracy the best 
performance is obtained with $V_4$ for two out of the four datasets and with $V_5$ 
for the other two; however the differences between them are relatively small. 
It should also be noted that $V_3$ gives the second best accuracy for three out of 
the four datasets. Overall, if we average the performance of each metric over the 
four datasets, $V_5$ gives the best accuracy, $V_3$ gives the lowest cross-entropy error
(which is lower than that of the traditional NN classifier), $V_5$ gives the lowest Brier 
score and $V_4$ gives the best reliability. This shows the advantage of using the four new 
Venn taxonomy definitions proposed in this work. The superior performance of the 
proposed definitions is due to the more effective partitioning of the examples into 
smaller categories and consequently 
the higher similarity between the examples in each category, which results in the 
probabilistic outputs of the VP being more accurate. As the performance of the 
proposed taxonomies varies across tasks, the most suitable taxonomy and the 
best threshold value~$\theta$ for a particular task can be chosen by experimentation 
on the available training examples.

\section{Conclusions}\label{sec:Conc}

This paper presented five Venn Predictors based on Neural Networks. Unlike the 
traditional NN classifiers the proposed methods produce probability intervals 
for each of their predictions, which are well-calibrated under the general i.i.d. assumption. 
The experiments performed in the on-line setting demonstrated the well-calibratedness 
of the probability intervals produced by the NN-VPs and their superiority 
over the single probabilities produced by traditional NNs, which can be significantly 
different from the observed frequencies. Moreover, the comparison performed 
in the batch setting showed that even when one discards the interval 
information produced by the NN-VPs by taking the mean of their multiprobability 
predictions, these are still much more reliable in most cases than the 
probabilities produced by traditional NNs. Lastly, the batch setting 
experiments also showed that NN-VPs are more accurate.

An important drawback of the VP approach is its computational inefficiency, 
which is a result of its transductive nature. Consequently an immediate future direction 
of this work is the development of an inductive VP approach based on the Inductive 
Conformal Prediction idea \cite{papa:icpnn} in order to overcome this computational inefficiency 
problem when dealing with large datasets. In addition, the application of 
VP to the problem of osteoporosis risk assessment is currently being studied. 
Its application to other challenging real world problems and the evaluation of its 
results is also of great interest.

\subsubsection*{Acknowledgments.} The author is grateful to
Professors V. Vovk and A. Gammerman for useful discussions. 
This work was supported by the European Regional Development 
Fund and the Cyprus Government through the Cyprus Research 
Promotion Foundation ``DESMI 2009-2010'' research contract 
TPE/ORIZO/0609(BIE)/24 (``Development of 
New Venn Prediction Methods for Osteoporosis Risk Assessment'').

\appendix

\section{On-line Experiments with $V_2, V_3, V_4$ and $V_5$}
\label{app1}

\begin{figure}
	\centering
		\subfloat[$V_2$]{\includegraphics[trim = 1mm 3mm 2mm 0mm, clip, width=6cm]{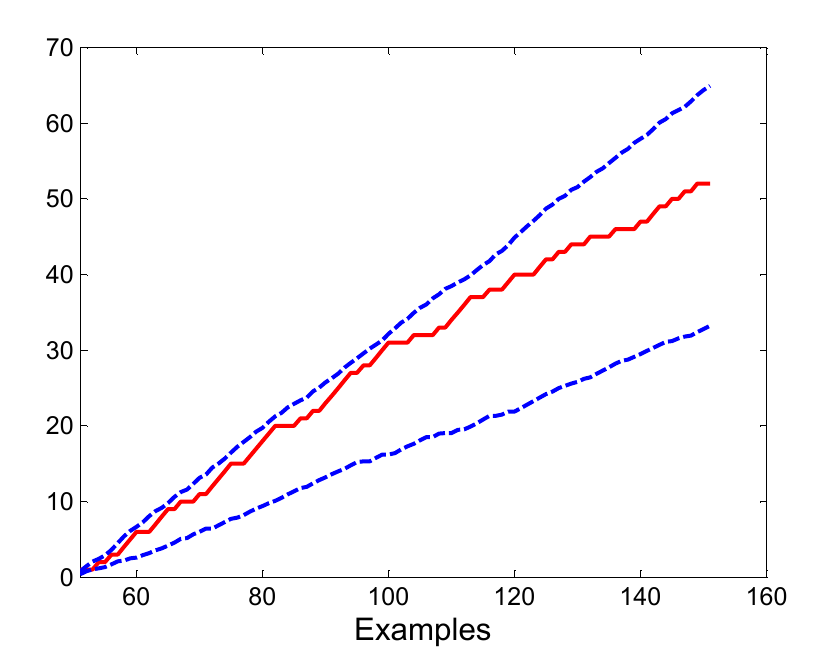}}
		\subfloat[$V_3$]{\includegraphics[trim = 2mm 3mm 2mm 0mm, clip, width=6cm]{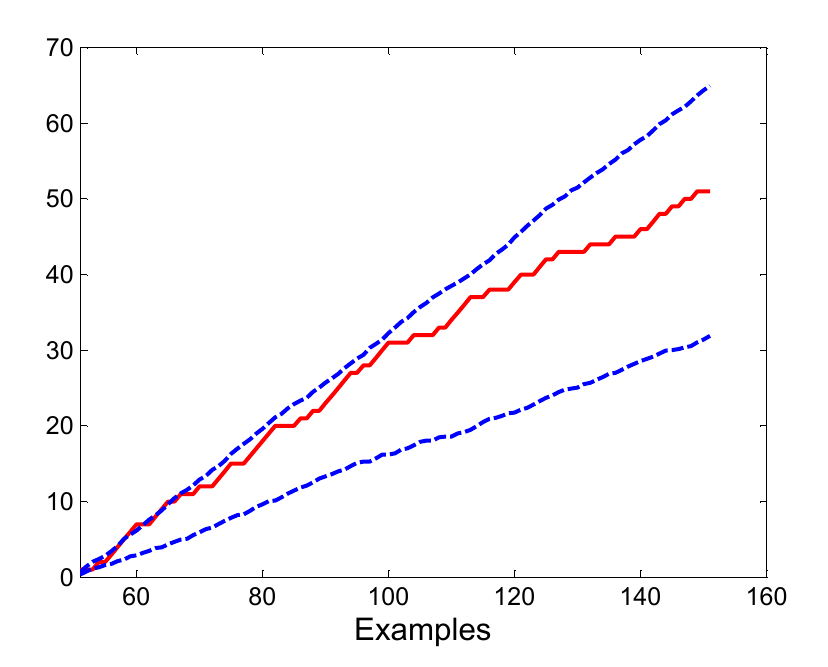}}\\
		\subfloat[$V_4$]{\includegraphics[trim = 2mm 3mm 2mm 0mm, clip, width=6cm]{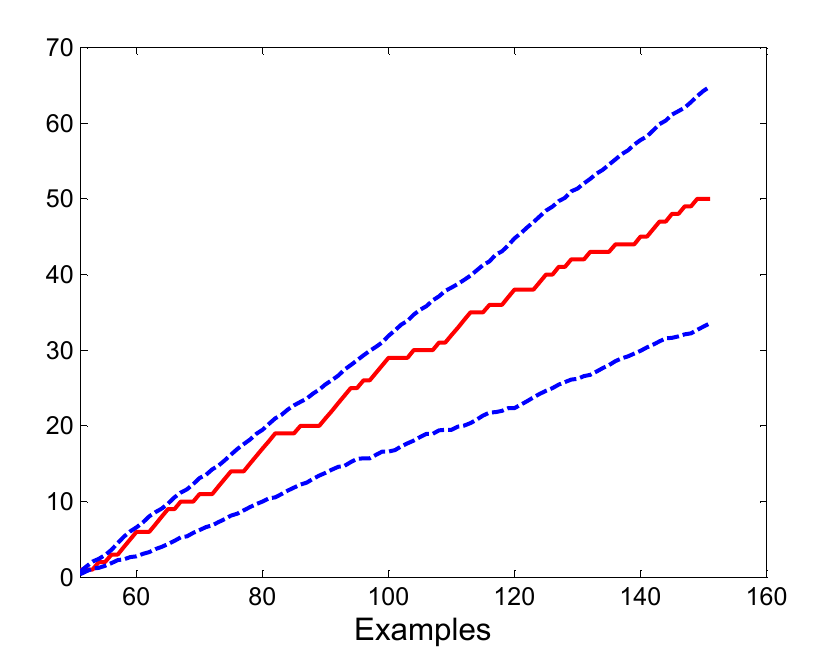}}
		\subfloat[$V_5$]{\includegraphics[trim = 2mm 3mm 2mm 0mm, clip, width=6cm]{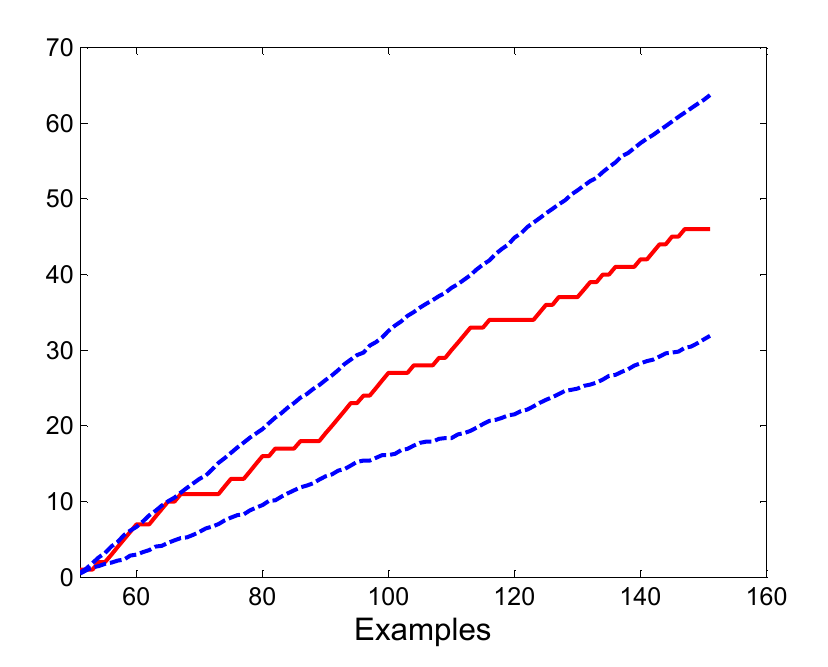}}
\caption{On-line performance of NN-VP with $V_2, V_3, V_4$ and $V_5$ on the TA Evaluation dataset. 
         Each plot shows the cumulative number of errors $E_n$ with a solid line and the cumulative lower and upper 
         error probability curves $LEP_n$ and $UEP_n$ with dashed lines.}
\label{fig:onlineVennTAE}
\end{figure}

\begin{figure}
	\centering
		\subfloat[$V_2$]{\includegraphics[trim = 1mm 3mm 2mm 0mm, clip, width=6cm]{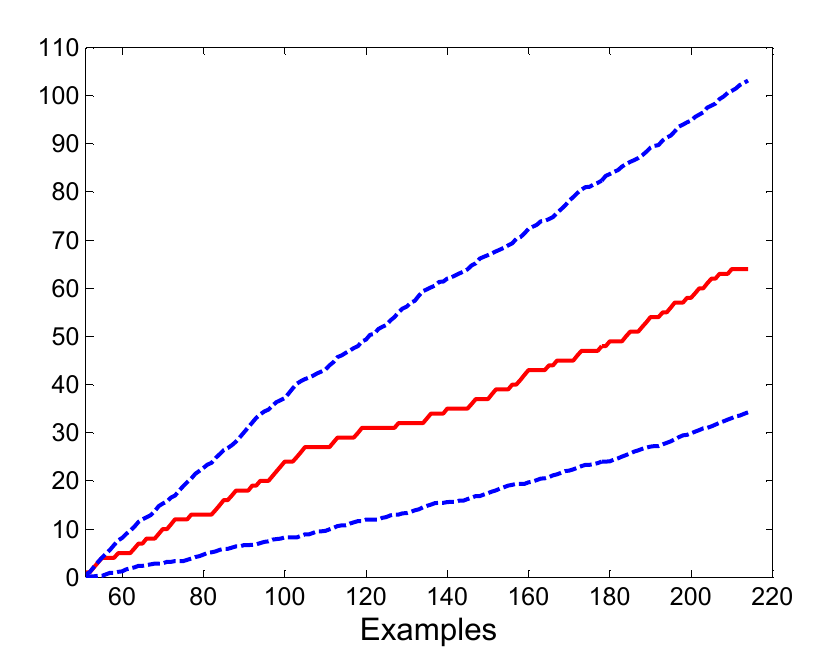}}
		\subfloat[$V_3$]{\includegraphics[trim = 2mm 3mm 2mm 0mm, clip, width=6cm]{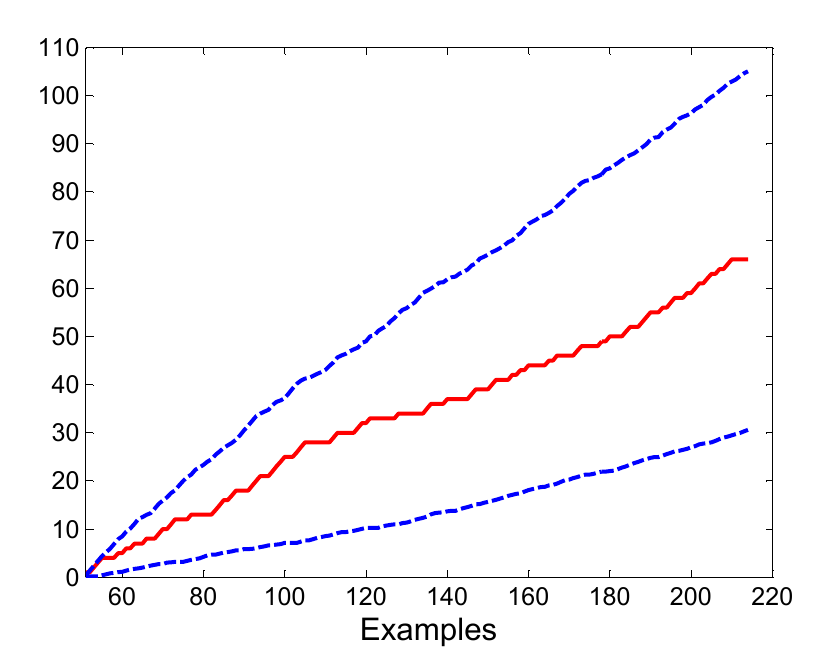}}\\
		\subfloat[$V_4$]{\includegraphics[trim = 2mm 3mm 2mm 0mm, clip, width=6cm]{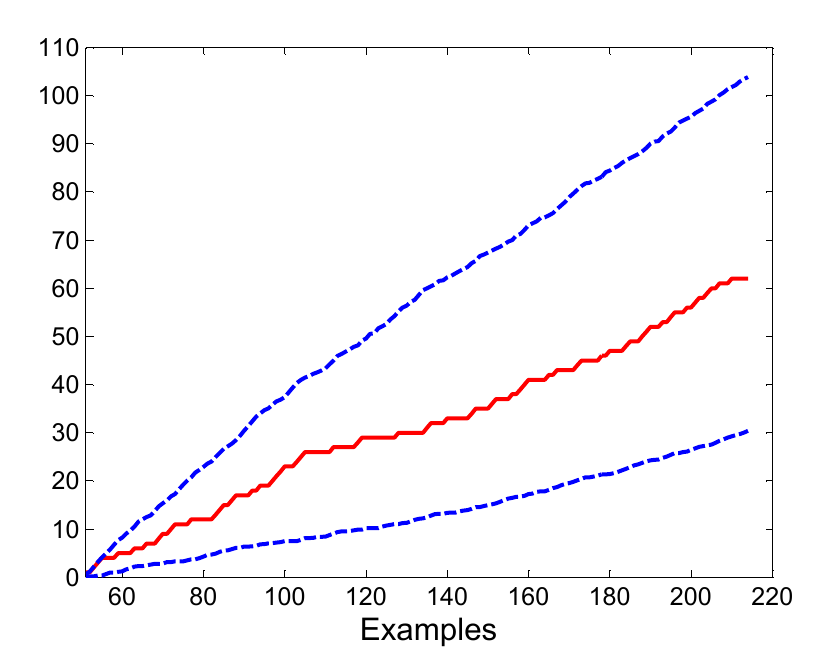}}
		\subfloat[$V_5$]{\includegraphics[trim = 2mm 3mm 2mm 0mm, clip, width=6cm]{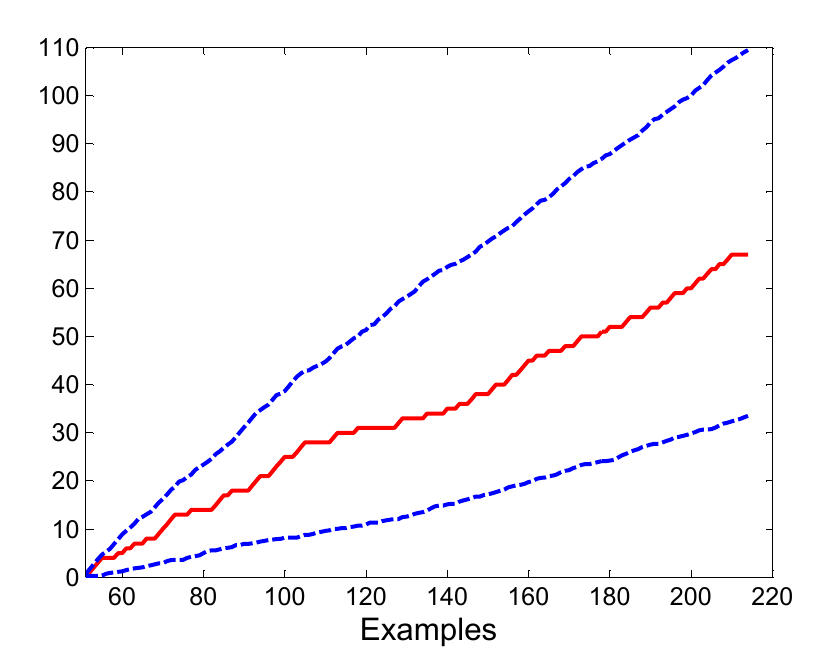}}
\caption{On-line performance of NN-VP with $V_2, V_3, V_4$ and $V_5$ on the Glass Identification dataset. 
         Each plot shows the cumulative number of errors $E_n$ with a solid line and the cumulative lower and upper 
         error probability curves $LEP_n$ and $UEP_n$ with dashed lines.}
\label{fig:onlineVennGlass}
\end{figure}

\begin{figure}
	\centering
		\subfloat[$V_2$]{\includegraphics[trim = 1mm 3mm 2mm 0mm, clip, width=6cm]{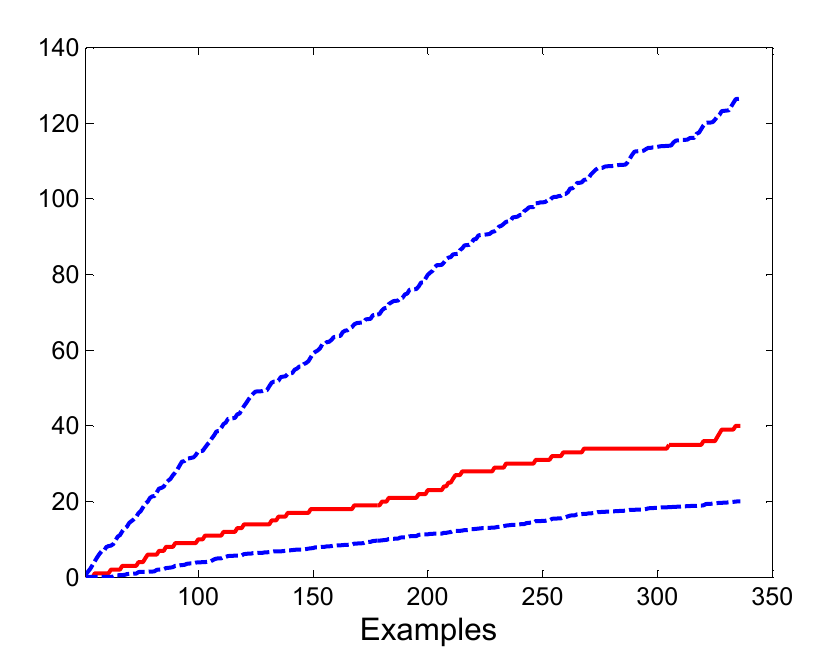}}
		\subfloat[$V_3$]{\includegraphics[trim = 2mm 3mm 2mm 0mm, clip, width=6cm]{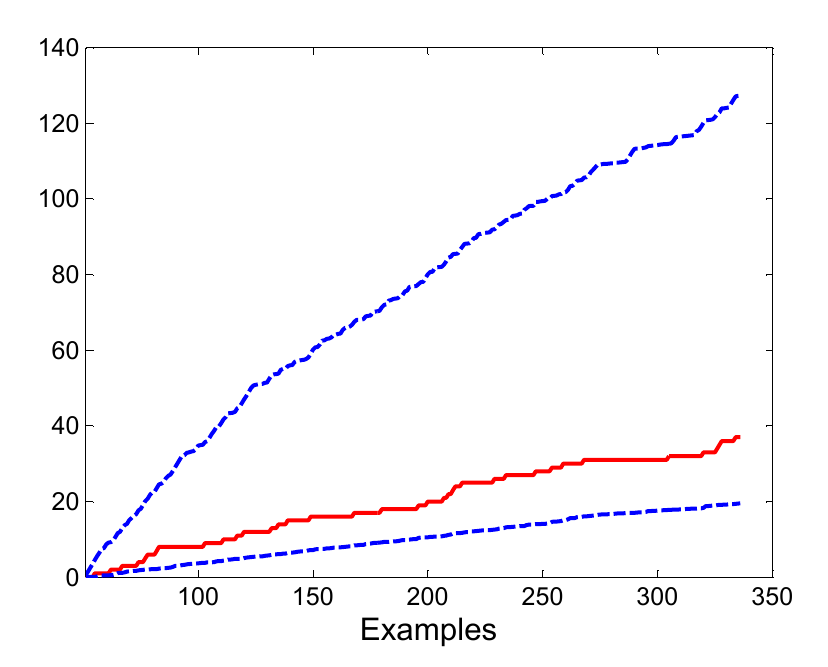}}\\
		\subfloat[$V_4$]{\includegraphics[trim = 2mm 3mm 2mm 0mm, clip, width=6cm]{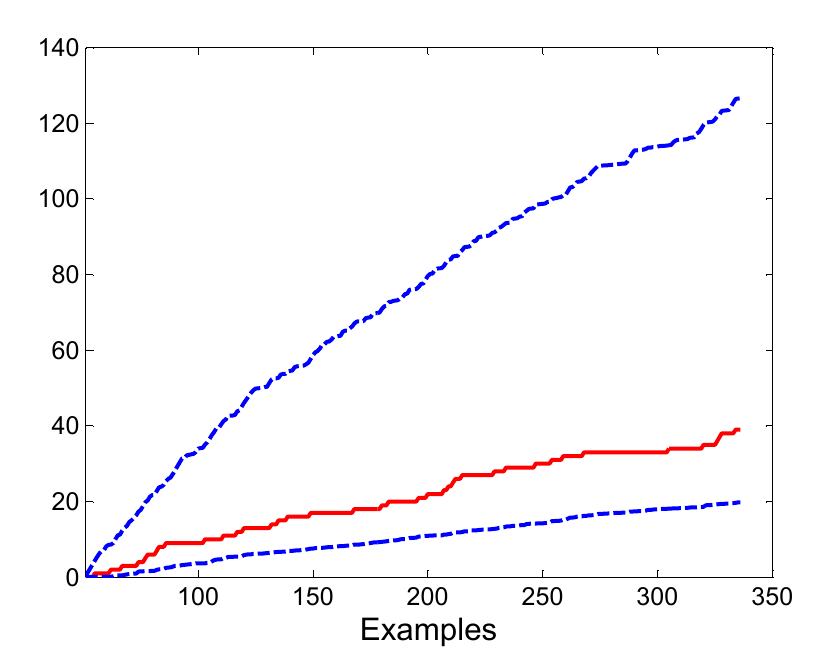}}
		\subfloat[$V_5$]{\includegraphics[trim = 2mm 3mm 2mm 0mm, clip, width=6cm]{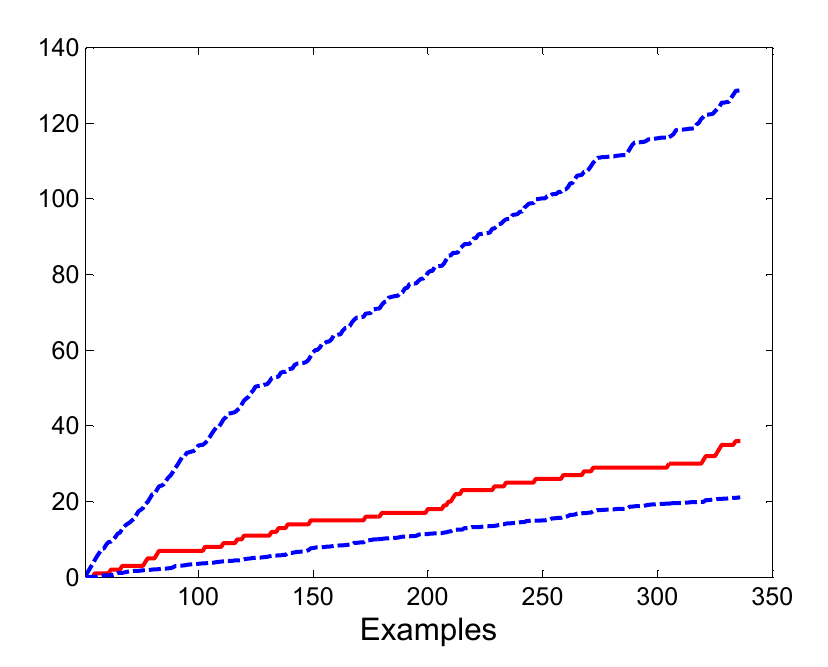}}
\caption{On-line performance of NN-VP with $V_2, V_3, V_4$ and $V_5$ on the Ecoli dataset. 
         Each plot shows the cumulative number of errors $E_n$ with a solid line and the cumulative lower and upper 
         error probability curves $LEP_n$ and $UEP_n$ with dashed lines.}
\label{fig:onlineVennEcoli}
\end{figure}

\begin{figure}
	\centering
		\subfloat[$V_2$]{\includegraphics[trim = 1mm 3mm 2mm 0mm, clip, width=6cm]{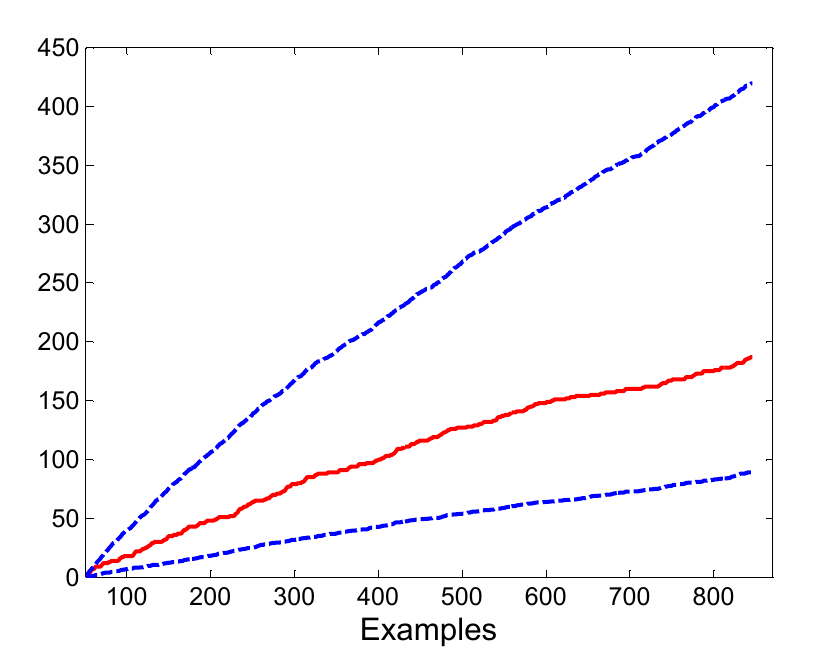}}
		\subfloat[$V_3$]{\includegraphics[trim = 2mm 3mm 2mm 0mm, clip, width=6cm]{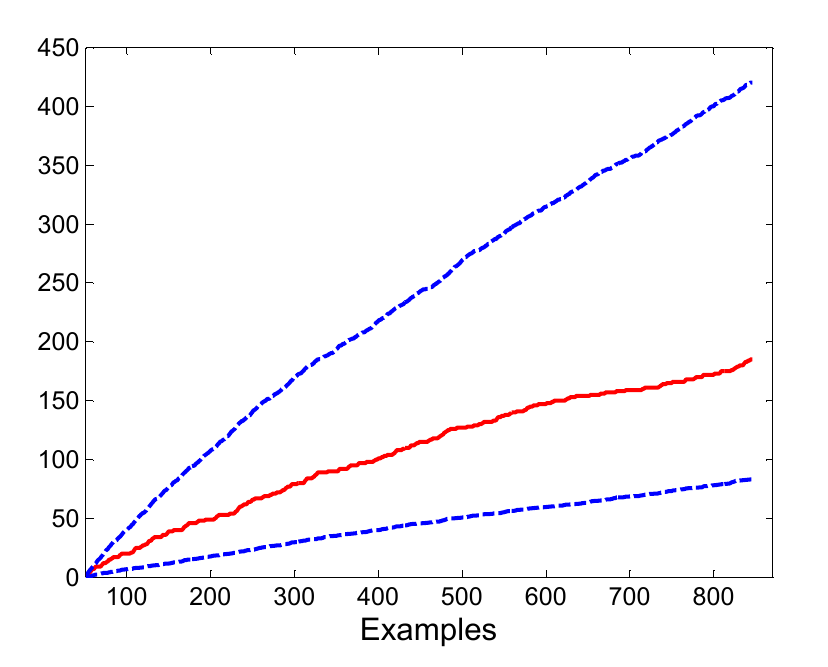}}\\
		\subfloat[$V_4$]{\includegraphics[trim = 2mm 3mm 2mm 0mm, clip, width=6cm]{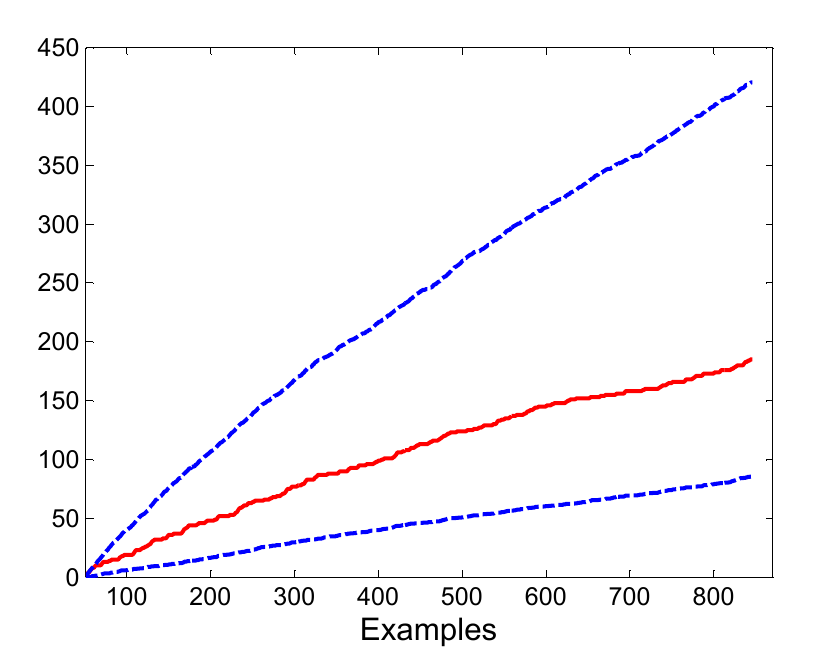}}
		\subfloat[$V_5$]{\includegraphics[trim = 2mm 3mm 2mm 0mm, clip, width=6cm]{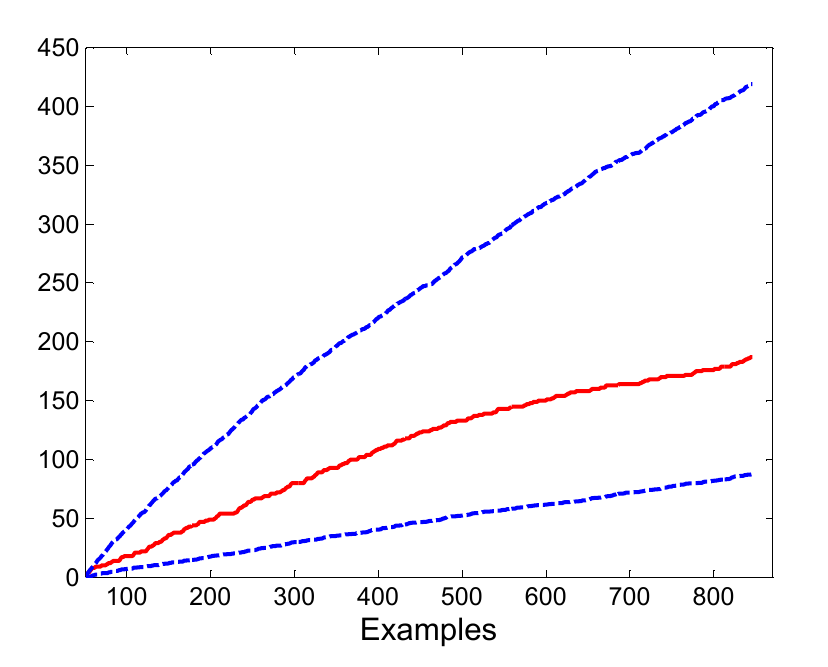}}
\caption{On-line performance of NN-VP with $V_2, V_3, V_4$ and $V_5$ on the Vehicle Silhouettes dataset. 
         Each plot shows the cumulative number of errors $E_n$ with a solid line and the cumulative lower and upper 
         error probability curves $LEP_n$ and $UEP_n$ with dashed lines.}
\label{fig:onlineVennVehicle}
\end{figure}

This Appendix presents in Figures~\ref{fig:onlineVennTAE} to~\ref{fig:onlineVennVehicle} the plots 
produced when applying the NN-VP with taxonomies $V_2, V_3, V_4$ and $V_5$ in the on-line mode to 
the four datasets.
Like in the plots obtained with taxonomy $V_1$, the cumulative errors are always included inside 
the cumulative upper and lower error probability curves produced by the NN-VP, which demonstrates 
that the probability intervals produced by NN-VP are well-calibrated.

\clearpage

\bibliographystyle{model1-num-names}
\bibliography{NNet-VP}







\end{document}